\documentclass{article}

\usepackage[accepted]{icml2018}

\usepackage[T1]{fontenc}

\usepackage{tgtermes}
\usepackage{amsmath}
\usepackage{scalefnt,letltxmacro}
\LetLtxMacro{\oldtextsc}{\textsc}
\renewcommand{\textsc}[1]{\oldtextsc{\scalefont{1.10}#1}}
\usepackage[scaled=0.92]{PTSans}

\usepackage[usenames,dvipsnames]{xcolor}
\definecolor{shadecolor}{gray}{0.9}

\usepackage[english]{babel}
\usepackage[parfill]{parskip}
\usepackage{afterpage}
\usepackage{framed}
\usepackage{nicefrac}

\usepackage{lineno}

\usepackage{ragged2e}

\usepackage{graphicx}
\usepackage[labelfont=bf]{caption}
\usepackage{subfig}

\usepackage{booktabs}
\usepackage{tabu}

\usepackage{natbib}

\usepackage{listings}
\usepackage{fancyvrb}
\fvset{fontsize=\normalsize}

\usepackage[colorlinks,linktoc=all]{hyperref}
\usepackage[all]{hypcap}
\hypersetup{citecolor=Violet}
\hypersetup{linkcolor=black}
\hypersetup{urlcolor=MidnightBlue}

\usepackage[nameinlink]{cleveref}

\usepackage
[acronym,smallcaps,nowarn,section,nogroupskip,nonumberlist]{glossaries}
\glsdisablehyper{}
\crefname{appendix}{supplement}{}

\usepackage{listings}
\lstdefinestyle{alp_style}{
    commentstyle=\color{OliveGreen},
    numberstyle=\tiny\color{black!60},
    stringstyle=\color{BrickRed},
    basicstyle=\ttfamily\scriptsize,
    breakatwhitespace=false,
    breaklines=true,
    captionpos=b,
    keepspaces=true,
    numbers=none,
    numbersep=5pt,
    showspaces=false,
    showstringspaces=false,
    showtabs=false,
    tabsize=2
}
\usepackage{bm}
\usepackage{amsmath,amssymb,dsfont}
\usepackage[usenames,dvipsnames]{xcolor}
\usepackage[ruled, vlined, linesnumbered]{algorithm2e}
\usepackage{wrapfig}
\usepackage{listings}

\icmltitlerunning{Quasi-Monte Carlo Variational Inference}

\newcommand{\RR}{\mathbb{R}}
\newcommand{\EE}{\mathbb{E}}
\newcommand{\KL}{\mathrm{KL}}

\DeclareMathOperator{\tr}{tr}

\newcommand{\SGD}{\textsc{sgd}}

\newcommand{\RQMC}{\textsc{rqmc}}
\newcommand{\MC}{\textsc{mc}}

\newcommand{\EL}{\textsc{elbo}}
\newcommand{\QMCVI}{\textsc{qmcvi}}
\newcommand{\CV}{\textsc{cv}}

\newtheorem{theorem}{Theorem}

\newcommand{\1}[1]{\mathds{1}\left\{#1\right\} }

\newcommand{\norm}[1]{\|#1\|^2_2 }
\newcommand{\Varr}[1]{\Var\left[#1\right]}

\DeclareMathOperator{\Var}{Var}
\newcommand{\uvec}{\mathbf{u}}
\newcommand{\OO}{\mathcal{O}}
\newcommand{\oo}{o}

\newcommand{\lambdavec}{\mathbf{\lambda}}
\newcommand{\lambdaveco}{\overline{\lambdavec}}
\newcommand{\lambdavectone}{\lambdavec_{t+1}}
\newcommand{\lambdavect}{\lambdavec_{t}}
\newcommand{\zvec}{\mathbf{z}}
\newcommand{\xvec}{\mathbf{x}}
\newcommand{\varepsilonvec}{\mathbf{\varepsilon}}
\newcommand{\muvec}{\mathbf{\mu}}

\newcommand{\ELBO}{\mathcal{L}}

\newcommand{\gradELBOestN}{\hat{g}_N}

\newcommand{\F}{F}

\newcommand{\FestN}{\hat{\F}_N}

\newcommand{\gradFestN}{\hat{g}_N}
\newcommand{\gradFestNt}{\hat{g}_{N_t}}

\newcommand{\sumnN}{\sum_{n=1}^N}
\newcommand{\sumtT}{\sum_{t=1}^T}
\newcommand{\limT}{\lim_{T\rightarrow\infty}}

\newcommand{\EEUNt}{\EE_{U_{N,t}}}

\newcommand{\UNt}{U_{N,t}}
\newcommand{\VarrUNt}[1]{\Var_{\UNt}\left[#1\right]}

\begin{document}

\twocolumn[
\icmltitle{Quasi-Monte Carlo Variational Inference}

\icmlsetsymbol{equal}{*}

\begin{icmlauthorlist}
\icmlauthor{Alexander Buchholz}{equal,pa}
\icmlauthor{Florian Wenzel}{equal,kl}
\icmlauthor{Stephan Mandt}{dis}

\end{icmlauthorlist}

\icmlaffiliation{pa}{ENSAE-CREST, Paris}
\icmlaffiliation{kl}{TU Kaiserslautern, Germany}
\icmlaffiliation{dis}{Los Angeles, CA, USA}

\icmlcorrespondingauthor{Alexander Buchholz}{alexander.buchholz@ensae.fr}
\icmlcorrespondingauthor{Florian Wenzel}{wenzelfl@hu-berlin.de}
\icmlcorrespondingauthor{Stephan Mandt}{stephan.mandt@gmail.com}

\icmlkeywords{Quasi-Monte Carlo, Black Box Variational Inference, Bayesian Approximate Inference}

\vskip 0.3in
]



\printAffiliationsAndNotice{\icmlEqualContribution} 

\begin{abstract}
Many machine learning problems involve Monte Carlo gradient estimators. 
As a prominent example, we focus on Monte Carlo variational inference (\textsc{mcvi}) in this paper. 
The performance of \textsc{mcvi} crucially depends on the variance of its stochastic gradients. We propose variance reduction by means of Quasi-Monte Carlo (\textsc{qmc}) sampling. \textsc{qmc} replaces $N$ i.i.d. samples from a uniform probability distribution by a deterministic sequence of samples of length $N$. This sequence
covers the underlying random variable space
more evenly than i.i.d. draws, reducing the variance of the gradient estimator. 
With our novel approach, both the score function and the reparameterization gradient estimators lead to much faster convergence.  
We also propose a new algorithm for Monte Carlo objectives, where
we operate with a constant learning rate and increase the number of \textsc{qmc} samples per iteration. We prove that this way, our algorithm can converge asymptotically at a faster rate than \textsc{sgd}.
We furthermore provide theoretical guarantees on \textsc{qmc} for Monte Carlo objectives that go beyond \textsc{mcvi}, and support our findings by several experiments on large-scale data sets from various domains.
\end{abstract}

\section{Introduction}

In many	situations in machine learning and statistics, we encounter objective functions which are expectations over continuous distributions. Among other examples, this situation occurs in reinforcement learning~\citep{sutton1998reinforcement} and variational inference~\citep{jordan1999introduction}. 
If the expectation cannot be computed in closed form, an approximation can often be obtained via Monte Carlo (\textsc{mc}) sampling from the underlying distribution. 
As most optimization procedures rely on the gradient of the objective, 
a \textsc{mc} gradient estimator has to be built by sampling from this distribution. The finite number of \textsc{mc}
samples per gradient step introduces noise. When averaging over multiple samples, the error in approximating the gradient
can be decreased, and thus its variance reduced. This guarantees stability and fast convergence of stochastic gradient descent (\textsc{sgd}).

Certain objective functions require a large number of \textsc{mc} samples per stochastic gradient step.
As a consequence, the algorithm gets slow. It is therefore desirable to obtain the same degree of variance reduction with
fewer samples. This paper proposes the idea of using Quasi-Monte Carlo (\textsc{qmc}) samples  instead of i.i.d. samples to achieve this goal.

A \textsc{qmc} sequence is a deterministic sequence which covers a hypercube $[0,1]^d$ more regularly than random samples.
When using a \textsc{qmc} sequence for Monte Carlo integration, the mean squared error (MSE) decreases asymptotically with the number of samples $N$ as ${\cal O}(N^{-2}(\log N)^{2d-2})$ \citep{leobacher2014introduction}.
In contrast, the naive \textsc{mc} integration error decreases as ${\cal O}(N^{-1})$. 
Since the cost of generating $N$ \textsc{qmc} samples is ${\cal O}(N \log N$), this
implies that a much smaller number of
operations per gradient step is required in order to achieve the same precision (provided that $N$ is large enough). Alternatively, we can achieve a larger variance reduction with the
same number of samples, allowing for larger gradient steps and therefore also faster convergence. 
This paper investigates the benefits of this approach both experimentally
and theoretically.

Our ideas apply in the context of Monte Carlo variational inference (\textsc{mcvi}),  a set of methods which make approximate Bayesian inference
scalable and easy to use. Variance reduction is an active area of research in this field. Our algorithm has the advantage of being very general; it can be easily implemented in existing
software packages such as STAN and Edward \cite{stan_JSSv076i01, tran2016edward}. In Appendix~\ref{appendix:rqmc_practice} we show how our  approach can be easily implemented in your existing code.

The main contributions are as follows:

\begin{itemize}
 \item We investigate the idea of using \textsc{qmc} sequences for Monte Carlo variational inference. 
  While the usage of \textsc{qmc} for \textsc{vi} has been	suggested in the outlook section of~\citet{DBLP:conf/aistats/RanganathGB14}, to our knowledge, we are the first to actually investigate this approach both theoretically and experimentally.
 \item We show that when using a randomized version of \textsc{qmc} (\textsc{rqmc}), the resulting stochastic gradient is unbiased and its variance is asymptotically reduced.
    We also show that when operating \textsc{sgd} with a constant learning rate, the stationary variance of the iterates is reduced by a factor of $N$, 
    allowing us to get closer to the optimum. 
\item We propose an algorithm which operates at a constant learning rate, but increases the number of \textsc{rqmc} samples over iterations. We prove that this algorithm has a better asymptotic convergence rate than \textsc{sgd}.
\item Based on three different experiments and for two popular types of gradient estimators we illustrate that our method allows us to
  train complex models several orders of magnitude faster than with standard \textsc{mcvi}.
\end{itemize}

Our paper is structured as follows. Section \ref{sec:related_work} reviews related work. Section \ref{sec:method} explains our method and exhibits our theoretical results. In Section \ref{sec:experiments} we describe our experiments and show comparisons to other existing approaches. Finally, Section \ref{sec:conclusion} concludes and lays out future research directions.

\section{Related Work} \label{sec:related_work}

\paragraph{Monte Carlo Variational Inference (MCVI)}

Since the introduction of the score function (or REINFORCE) gradient estimator for variational inference  \cite{paisley2012variational,DBLP:conf/aistats/RanganathGB14}, 
Monte Carlo variational inference has received an ever-growing attention, see \citet{zhang2017advances} for a recent review. The introduction of the gradient estimator made \textsc{vi} applicable to non-conjugate models but highly depends on the variance of the gradient estimator. Therefore various variance reduction techniques have been introduced; for example Rao-Blackwellization and control variates, see \citet{DBLP:conf/aistats/RanganathGB14} and importance sampling, see \citet{DBLP:conf/uai/RuizTB16, burda2015importance}. 

At the same time the work of \citet{DBLP:journals/corr/KingmaW13, DBLP:conf/icml/RezendeMW14} introduced reparameterization gradients for \textsc{mcvi}, which typically exhibits lower variance but are restricted to models where the variational family can be reparametrized via a differentiable mapping. 
In this sense \textsc{mcvi} based on score function gradient estimators is more general but training the algorithm is more difficult. A unifying view is provided by \citet{ruiz2016generalized}. 
\citet{DBLP:conf/nips/MillerFDA17} introduce a modification of the reparametrized version, but relies itself on assumptions on the underlying variational family. 
\citet{DBLP:conf/nips/RoederWD17} propose a lower variance gradient estimator by omitting a term of the \textsc{elbo}. The idea of using \textsc{qmc} in order to reduce the variance has been suggested by \citet{DBLP:conf/aistats/RanganathGB14} and \citet{DBLP:conf/uai/RuizTB16} and used for a specific model by \citet{vb_intractablelikelihood}, but without a focus on analyzing or benchmarking the method.

\paragraph{Quasi-Monte Carlo and Stochastic Optimization}
Besides the generation of random samples for approximating posterior distributions \citep{robert2013monte}, 
Monte Carlo methods are used for calculating expectations of intractable integrals via the law of large numbers. 
The error of the integration with random samples goes to zero at a rate of $\mathcal{O}(N^{-1})$ in terms of the MSE. For practical application this rate can be too slow. 
Faster rates of convergence in reasonable dimensions can be obtained by replacing the randomness by a deterministic sequence, also called Quasi-Monte Carlo. 

Compared to Monte Carlo and for sufficiently regular functions, \textsc{qmc} reaches a faster rate of convergence of the approximation error of an integral.
\citet{niederreiter1992random, l2005recent, leobacher2014introduction, dick_kuo_sloan_2013} provide excellent reviews on this topic. 
From a theoretical point of view, the benefits of \textsc{qmc} vanish in very high dimensions. Nevertheless, the error bounds are often too pessimistic and in practice, gains are observed up to dimension $150$, see \citet{glasserman2013monte}. 

 \textsc{qmc} has frequently been used in financial applications \citep{glasserman2013monte, joy1996quasi, lemieux2001use}. 
In statistics, some applications include particle filtering \citep{gerber2015sequential}, approximate Bayesian computation \citep{buchholz2017improving}, control functionals \citep{oates2016control} and Bayesian optimal design \citep{drovandi2018}.  \citet{yang2014quasi} used  \textsc{qmc} in the context of large scale kernel methods.

Stochastic optimization has been pioneered by the work of \citet{robbins1951stochastic}. 
As stochastic gradient descent suffers from noisy gradients, various approaches for reducing the variance and adapting the step size have been introduced \citep{johnson2013accelerating, kingma2014adam, defazio2014saga, duchi2011adaptive,zhang2017determinantal}. Extensive theoretical results on the convergence of stochastic gradients algorithms are provided by \citet{bachmoulines2011}. \citet{mandt2017stochastic} interpreted stochastic gradient descent with constant learning rates as approximate Bayesian inference. Some recent reviews are for example \citet{bottou2016optimization, nesterov2013introductory}. Naturally, concepts from \textsc{qmc} can be beneficial to stochastic optimization. Contributions on exploiting this idea are e.g. \citet{gerber2017improving} and \citet{drew2006quasi}.

\section{Quasi-Monte Carlo Variational Inference} \label{sec:method}
In this Section, we introduce Quasi-Monte Carlo Variational Inference (\textsc{qmcvi}),  using randomized \textsc{qmc} (\textsc{rqmc}) for variational inference. We review \textsc{mcvi} in Section~\ref{sec:mcvi}. \textsc{rqmc} and the details of our algorithm are exposed in Section~\ref{sec:qmcvi}. Theoretical results are given in Section~\ref{sec:theory}.

\subsection{Background: Monte Carlo Variational Inference}
\label{sec:mcvi}

Variational inference (\textsc{vi}) is key to modern probabilistic modeling and Bayesian deep learning~\citep{jordan1999introduction,blei2017variational,zhang2017advances}. In Bayesian inference, the object of interest is a posterior distribution of latent variables $\zvec$ given observations $\xvec$. \textsc{vi} approximates Bayesian inference by an optimization problem which we can solve by (stochastic) gradient ascent~\citep{jordan1999introduction,hoffman2013stochastic}. 

In more detail, \textsc{vi} builds a tractable approximation of the posterior $p(\zvec |\xvec)$ by minimizing the $\KL$-divergence between a variational family $q(\zvec|\lambdavec)$, parametrized by free parameters $\lambdavec \in \RR^d$, and $p(\zvec|\xvec)$. 
This is equivalent to maximizing the so-called evidence lower bound (\textsc{elbo}):
\begin{eqnarray} \label{eq:elbo}
\ELBO(\lambdavec) = \EE_{q(\zvec|\lambdavec)}[ \log p(\xvec,\zvec) - \log q(\zvec|\lambdavec)].
\end{eqnarray}
In classical variational inference, the expectations involved in \eqref{eq:elbo} are carried out analytically~\citep{jordan1999introduction}. However, this is only possible for the fairly restricted class of so-called conditionally conjugate exponential family models~\citep{hoffman2013stochastic}. More recently, black-box variational methods have gained momentum, which make the analytical evaluation of these expectation redundant, and which shall be considered in this paper.

Maximizing the objective \eqref{eq:elbo} is often based on a gradient ascent scheme. 
However, a direct differentiation of the objective \eqref{eq:elbo} with respect to $\lambdavec$ is not possible, as the measure of the expectation depends on this parameter. The two major approaches for overcoming this issue are the score function estimator and the reparameterization estimator.

\paragraph{Score Function Gradient}
The score function gradient (also called REINFORCE gradient)~\citep{DBLP:conf/aistats/RanganathGB14}
expresses the gradient as expectation with respect to $q(\zvec|\lambdavec)$ and is given by
\begin{multline} \label{eq:grad_reinforce}
\nabla_\lambdavec \ELBO(\lambdavec) \\ = \EE_{q(\zvec|\lambdavec)}[\nabla_\lambdavec \log q(\zvec|\lambdavec) \left(\log p(\xvec,\zvec) - \log q(\zvec|\lambdavec)\right)].
\end{multline}
The gradient estimator is obtained by approximating the expectation with independent samples from the variational distribution $q(\zvec|\lambdavec)$.
This estimator applies to continuous and discrete variational distributions. 

\paragraph{Reparameterization Gradient}
The second approach is based on the reparametrization trick~\citep{DBLP:journals/corr/KingmaW13}, where the distribution over $\zvec$ is expressed as a deterministic transformation of another distribution over a noise variable $\varepsilonvec$, hence  $\zvec = g_\lambdavec(\varepsilon)$ where $\varepsilonvec \sim p(\varepsilonvec)$. Using the reparameterization trick,  the \EL{} is expressed as expectation with respect to $p(\varepsilon)$ and the derivative is moved inside the expectation:
\begin{multline} \label{eq:grad_repara}
\nabla_\lambdavec \ELBO(\lambdavec) \\ = \EE_{p(\varepsilonvec)}[\nabla_\lambdavec \log p(\xvec,g_\lambdavec(\varepsilonvec)) - \nabla_\lambdavec  \log q(g_\lambdavec(\varepsilonvec)|\lambdavec)].
\end{multline}
The expectation is approximated using a \MC~sum of independent samples from $p(\varepsilon$). In its basic form, the estimator is restricted to distributions over continuous variables.


\paragraph{MCVI}
In the general setup of \textsc{mcvi} considered here, the gradient of the \textsc{elbo} is represented as an expectation $\nabla_\lambdavec \ELBO(\lambdavec) \\ = \EE[g_{\tilde \zvec}(\lambda)]$ over a random variable $\tilde \zvec$. For the score function estimator we choose $g$ according to Equation~\eqref{eq:grad_reinforce} with $\tilde \zvec = \zvec$ and for the reparameterization gradient according to Equation~\eqref{eq:grad_repara} with $\tilde \zvec = \varepsilonvec$, respectively.
This allows us to obtain a stochastic estimator of the gradient by an average over a finite sample
$\{\tilde \zvec_1, \cdots, \tilde \zvec_N\}$ as  $\hat{g}_N(\lambdavec_t) = (1/N) \sum_{i=1}^N g_{\tilde \zvec_i}( \lambdavec_t)$.
This way,
the \textsc{elbo} can be optimized by stochastic optimization. This is achieved by iterating the \SGD~updates with decreasing step sizes $\alpha_t$:
\begin{equation} \label{eq:sgd_update_elbo}
\lambdavec_{t+1} = \lambdavec_{t} + \alpha_t \gradELBOestN(\lambdavec_{t}).
\end{equation}
The convergence of the gradient ascent scheme in \eqref{eq:sgd_update_elbo} tends to be slow when gradient estimators have a high variance.  
Therefore, various approaches for reducing the variance of both gradient estimators exist; e.g. control variates (\CV{}), Rao-Blackwellization and importance sampling. However these variance reduction techniques do not improve the $\OO(N^{-1})$ rate of the MSE of the estimator, except under some restrictive conditions \citep{oates2017control}. Moreover, the variance reduction schemes must often be tailored to the problem at hand.

\subsection{Quasi-Monte Carlo Variational Inference}
\label{sec:qmcvi}
\paragraph{Quasi Monte Carlo }
Low discrepancy sequences, also called \textsc{qmc} sequences, are used for integrating a function $\psi$ over the $[0,1]^d$ hypercube. When using standard i.i.d. samples on $[0,1]^d$, the error of the approximation is $\OO(N^{-1})$. \textsc{qmc} achieves a rate of convergence in terms of the MSE of $\OO\left( N^{-2}(\log N)^{2d-2} \right)$ if $\psi$ is sufficiently regular \citep{leobacher2014introduction}. This is achieved by a deterministic sequence that covers $[0,1]^d$ more evenly. 

On a high level, \textsc{qmc} sequences are constructed such that the number of points that fall in a rectangular volume is proportional to the volume. This idea is closely linked to stratification. Halton sequences e.g. are constructed using coprime numbers \citep{halton1964algorithm}. Sobol sequences are based on the reflected binary code \citep{antonov1979economic}. The exact construction of \textsc{qmc} sequences is quite involved and we refer to \citet{niederreiter1992random, leobacher2014introduction, dick_kuo_sloan_2013} for more details.

The approximation error of \textsc{qmc} increases with the dimension, and it is difficult to quantify. 
Carefully reintroducing randomness while preserving the structure of the sequence leads to randomized \textsc{qmc}. 
\textsc{rqmc} sequences are unbiased and the error can be assessed by repeated simulation. 
Moreover, under slightly stronger regularity conditions on $F$ we can achieve rates of convergence of $\OO(N^{-2})$ \citep{gerber2015integration}. 
For illustration purposes, we show different sequences in Figure \ref{fig:uniform_sequences}. In Appendix~\ref{appendix:info_qmc} we provide more technical details. 

\textsc{qmc} or \textsc{rqmc} can be used for integration with respect to arbitrary distributions by transforming 
the initial sequence on $[0,1]^d$ via a transformation $\Gamma$ to the distribution of interest. Constructing the sequence typically costs $\OO(N \log N)$ \citep{gerber2015sequential}.

\begin{figure}
    \centering
    \includegraphics[width=0.49\textwidth]{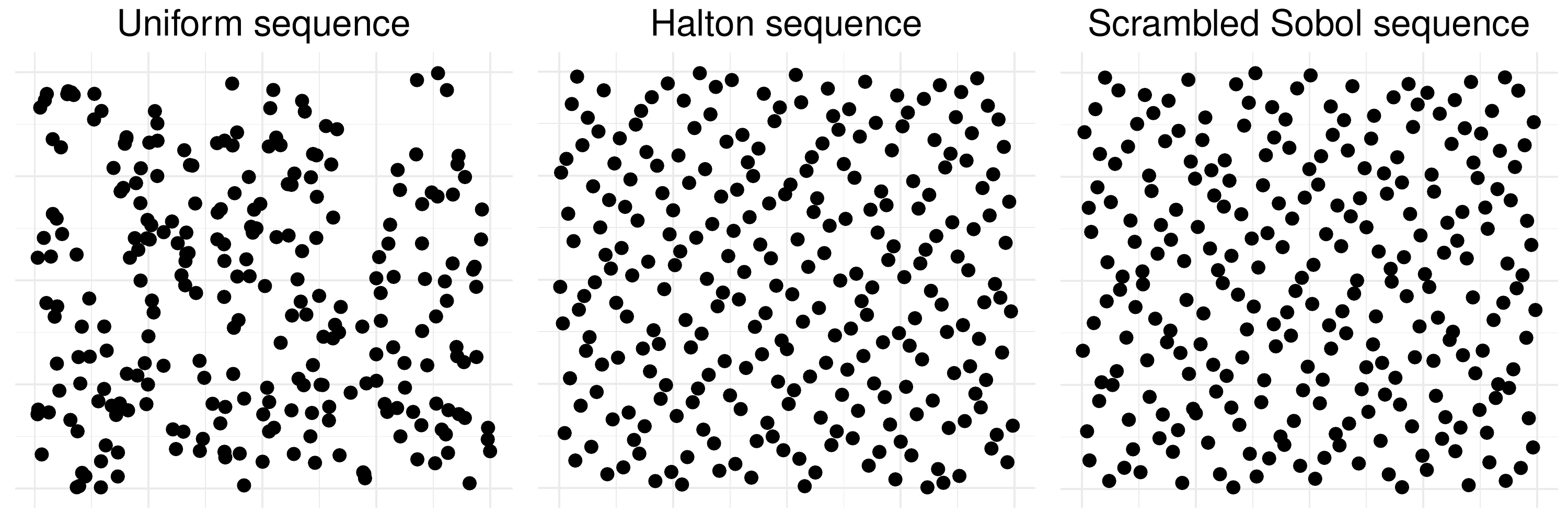}
    \caption{\textsc{mc} (left), \textsc{qmc} (center) and \textsc{rqmc} (right) sequences of length $N=256$ on $[0,1]^2$. \textsc{qmc} and \textsc{rqmc} tend to cover the target space more evenly.}
    \label{fig:uniform_sequences}
\end{figure}

\paragraph{QMC and VI}
We suggest to replace $N$ independent  \textsc{mc} samples for computing $\gradELBOestN(\lambdavec_{t})$ by an \textsc{rqmc} sequence of the same length. With our approach, the variance of the gradient estimators becomes $\OO(N^{-2})$, and the costs for creating the sequence is $\OO(N \log N)$.
The incorporation of \textsc{rqmc} in \textsc{vi} is straightforward: instead of sampling
$\tilde \zvec$
as independent \textsc{mc} samples, we generate a uniform \textsc{rqmc} sequence $\uvec_1, \cdots, \uvec_N$
and transform this sequence via a mapping $\Gamma$ to the original random variable $\tilde \zvec = \Gamma(\uvec)$. 
Using this transformation we obtain the \RQMC{} gradient estimator
\begin{equation}
\label{eq:reparameterization_fct}
    \hat{g}_N(\lambdavec_t) = (1/N) \sum_{i=1}^N g_{\Gamma(\uvec_i)}(\lambdavec).
\end{equation}
%
From a theoretical perspective, the function 
$\uvec \mapsto g_{\Gamma(\uvec)}(\lambdavec)$ has to be sufficiently smooth for all $\lambdavec$. 
%
For commonly used variational families this transformation is readily available. 
Although evaluating these transforms adds computational overhead, we found this cost negligible in practice.
For example, in order to sample from a multivariate Gaussian $\zvec_n \sim \mathcal{N}(\muvec, \Sigma)$, we generate an \textsc{rqmc} squence $\uvec_n$ and apply the transformation $\zvec_n = \Phi^{-1}(\uvec_n)\Sigma^{1/2} + \muvec$, where $\Sigma^{1/2}$ is the Cholesky decomposition of $\Sigma$ and $\Phi^{-1}$ is the component-wise inverse cdf of a standard normal distribution. Similar procedures are easily obtained for exponential, Gamma, and other distributions that belong to the exponential family. Algorithm \ref{algo:algo_qmcvi} summarizes the procedure. 

    \begin{algorithm}

        \KwIn{Data $\xvec$, model $p(\xvec,\zvec)$, variational family $q(\zvec|\lambda)$}
        \KwResult{Variational parameters $\lambdavec^*$}
        \While{not converged}{
            Generate uniform \textsc{rqmc} sequence $\uvec_{1:N}$\\
            Transform the sequence via $\Gamma$\\
            Estimate the gradient $\hat g_N(\lambdavec_{t}) = \frac{1}{N}\sum_{i=1}^N g_{\Gamma(\uvec_{i})}(\lambdavec_{t})$\\
            Update $\lambdavec_{t+1} = \lambdavec_{t} + \alpha_t$ $\hat g_N(\lambdavec_{t})$
        }
    \caption{Quasi-Monte Carlo Variational Inference}
    \label{algo:algo_qmcvi}
    \end{algorithm}

\textsc{rqmc} samples can be generated via standard packages such as \texttt{randtoolbox} \citep{randtoolbox}, available in \texttt{R}. Existing \textsc{mcvi} algorithms are adapted by replacing the random variable sampler by an \textsc{rqmc} version. 
Our approach reduces the variance in \textsc{mcvi} and applies in particular to
the reparametrization gradient estimator and the score function estimator.
\textsc{rqmc} can in principle be combined with additional variance reduction techniques such as \textsc{cv}, but care must be taken as the optimal \textsc{cv} for \textsc{rqmc} are not the same as for \textsc{mc} \citep{hickernell2005control}. 
 

\subsection{Theoretical Properties of QMCVI} \label{sec:theory}

In what follows we give a theoretical analysis of using \textsc{rqmc} in stochastic optimization. 
Our results apply in particular to \textsc{vi} but are more general.

\textsc{qmcvi} leads to faster convergence in combination with Adam \citep{kingma2014adam} or Adagrad \citep{duchi2011adaptive}, as we will show empirically in Section~\ref{sec:experiments}. Our  analysis, presented in this section, underlines this statement for the simple case of \textsc{sgd} with fixed step size in the Lipschitz continuous (Theorem \ref{theorem:lipschitz_functions}) and strongly convex case (Theorem \ref{theorem:strongly_convex_functions}). We show that for $N$ sufficiently large, \textsc{sgd} with \textsc{rqmc} samples reaches regions closer to the true optimizer of the \textsc{elbo}. Moreover, we obtain a faster convergence rate than \textsc{sgd} when using a fixed step size and increasing the sample size over iterations (Theorem \ref{theorem:noise_reduction}).

\paragraph{RQMC for Optimizing Monte Carlo Objectives}
We step back from black box variational inference and consider the more general setup of optimizing Monte Carlo objectives. Our goal is to minimize a function $F(\lambda)$, where 
the optimizer has only access to a noisy, unbiased version $\FestN (\lambdavec)$, with $\EE [\FestN (\lambdavec)] = \F(\lambdavec)$
and access to an unbiased noisy estimator of the gradients $\gradFestN(\lambdavec)$, with $\EE[\gradFestN(\lambdavec)] = \nabla \F(\lambdavec)$. The optimum of $F(\lambda)$ is $\lambdavec^\star$.


We furthermore assume that the gradient estimator $\gradFestN(\lambdavec)$ has the form as in Eq.~\ref{eq:reparameterization_fct}, where $\Gamma$ is a reparameterization function that converts uniform samples from the hypercube into samples from the target distribution.
In this paper, $\uvec_1, \cdots, \uvec_N$ is an \textsc{rqmc} sequence.

In the following theoretical analysis, we focus on \textsc{sgd} with a constant learning rate $\alpha$. 
The optimal value $\lambdavec^\star$ is approximated by \textsc{sgd} using the update rule
\begin{equation} \label{eq:sgd_update}
\lambdavec_{t+1} = \lambdavec_{t} - \alpha \gradFestN(\lambdavec_{t}).
\end{equation}
Starting from $\lambdavec_{1}$ the procedure is iterated until $|\FestN(\lambdavec_{t})-\FestN(\lambdavec_{t+1})|\leq \epsilon$, for a small threshold $\epsilon$. The quality of the approximation $\lambdavec_{T} \approx \lambdavec^\star$ crucially depends on the variance of the estimator $\gradFestN$  \citep{johnson2013accelerating}.

Intuitively, the variance of $\gradFestN(\lambdavec)$ based on an \textsc{rqmc} sequence will be $\OO(N^{-2})$ and thus for $N$ large enough, the variance will be smaller than for the \textsc{mc} counterpart, that is $\OO(N^{-1})$.
This will be beneficial to the optimization procedure defined in \eqref{eq:sgd_update}. 
Our following theoretical results are based on standard proof techniques for stochastic approximation, see e.g. \citet{bottou2016optimization}. 

\paragraph{Stochastic Gradient Descent with Fixed Step Size}
In the case of functions with Lipschitz continuous derivatives, we obtain the following upper bound on the norm of the gradients. 
\begin{theorem} \label{theorem:lipschitz_functions}
Let $\F$ be a function with Lipschitz continuous derivatives, i.e. there exists $L>0$ s.t. $\forall \lambdavec, \overline{\lambdavec}$ $\norm{ \nabla \F(\lambdavec) - \nabla \F(\overline{\lambdavec}) } \leq L \norm{\lambdavec - \overline{\lambdavec} }$, let $U_N = \{\uvec_1, \cdots, \uvec_N\}$ be an \textsc{rqmc} sequence and let $ \forall \lambdavec$, $G: \uvec \mapsto g_{\Gamma(\uvec)}(\lambdavec)$ has cross partial derivatives of up to order $d$. Let the constant learning rate $\alpha < 2/L$ and let $\mu = 1 - \alpha L/2$. 
Then  
$ \forall \lambdavec, \tr \Var_{U_N}[ \gradFestN(\lambdavec)] \leq M_V \times r(N)$, where $M_V < \infty$ and $r(N)=\OO\left(N^{-2}\right)$ and
\begin{multline*} \label{eq:sum_lipschitz}
\frac{\sumtT \EE \norm{ \nabla \F(\lambdavec_t)}}{T} \\ \leq \frac{1}{2 \mu} \alpha L  M_V r(N) + 
 \frac{ \F(\lambdavec_1) -  \F(\lambdavec^\star) }{\alpha \mu T }, 
\end{multline*}
where $\lambdavect$ is iteratively defined in \eqref{eq:sgd_update}. Consequently, 
\begin{equation} \label{eq:limit_lipschitz}
\limT \frac{\sumtT \EE \norm{ \nabla \F(\lambdavec_t)}}{T} \leq \frac{1}{2 \mu} \alpha L  M_V r(N) .
\end{equation}
\end{theorem}
Equation \eqref{eq:limit_lipschitz} underlines the dependence of the sum of the norm of the gradients on the variance of the gradients. The better the gradients are estimated, the closer one gets to the optimum where the gradient vanishes. As the dependence on the sample size becomes $\OO\left(N^{-2}\right)$ for an \textsc{rqmc} sequence instead of $1/N$ for a \textsc{mc} sequence, the gradient is more precisely estimated for $N$ large enough. 

We now study the impact of a reduced variance on \textsc{sgd} with a fixed step size and strongly convex functions. We obtain an improved upper bound on the optimality gap.

\begin{theorem} \label{theorem:strongly_convex_functions}
Let $F$ have Lipschitz continuous derivatives and be a strongly convex function, i.e. there exists a constant $c>0$ s.t. $\forall \lambdavec, \overline{\lambdavec}$ $ \F(\overline{\lambdavec})  \geq \F(\lambdavec) + \nabla \F(\lambdavec)^T (\lambdavec -  \overline{\lambdavec}) + \frac{1}{2} c \norm{ \lambdavec - \overline{\lambdavec} }$, let $U_N = \{ \uvec_1, \cdots, \uvec_N \}$ be an \textsc{rqmc} sequence and let $ \forall \lambdavec, G: \uvec \mapsto g_{\Gamma(\uvec)}(\lambdavec)$ be as in Theorem \ref{theorem:lipschitz_functions}. Let the constant learning rate $\alpha < \frac{1}{2c}$ and $\alpha < \frac{2}{L}$. Then the expected optimality gap satisfies, $\forall t \geq 0$, 
\begin{multline*}
\EE[\F(\lambdavec_{t+1}) - \F(\lambdavec^\star)] \\ \leq  \left[ \left( \frac{\alpha^2 L}{2} - \alpha \right) 2c +1 \right ] \times \EE[\F_N(\lambdavec_{t})  - \F(\lambdavec^\star)]  \\+ \frac{1}{2} L \alpha^2 \left[ M_V r(N) \right].
\end{multline*}
Consequently, 
\begin{equation*} \label{eq:limit_strongly_convex}
\limT \EE[\F(\lambdavec_{T}) - \F(\lambdavec^\star)] \leq \frac{\alpha L }{4c - \alpha L c} \left[ M_V r(N) \right].
\end{equation*}
\end{theorem}
The previous result has the following interpretation. The expected optimality gap between the last iteration $\lambdavec_T$ and the true minimizer $\lambdavec^\star$ is upper bounded by the magnitude of the variance. The smaller this variance, the closer we get to $\lambdavec^\star$. Using \textsc{rqmc} we gain a factor $1/N$ in the bound. 

\paragraph{Increasing Sample Size Over Iterations}
While \textsc{sgd} with a fixed step size and a fixed number of samples per gradient step does not converge, convergence can be achieved when increasing the number of samples used for estimating the gradient over iterations. As an extension of Theorem \ref{theorem:strongly_convex_functions}, we show that a linear convergence is obtained while increasing the sample size at a slower rate than for \textsc{mc} sampling. 
\begin{theorem} \label{theorem:noise_reduction}
Assume the conditions of Theorem $\ref{theorem:strongly_convex_functions}$ with the modification $\alpha \leq \min \{1/c, 1/L \}$.
Let $ 1-\alpha c /2 <\xi^{2} = \frac{1}{\tau^2}<1$. Use an increasing sample size $N_t = \underline{N} + \lceil \tau^{t} \rceil $, where $\underline{N} < \infty$ is defined in Appendix~\ref{appendix:thm3}. Then $\forall t \in \mathbb{N}, \exists \hat{M}_V < \infty, $ 
$$
\tr \Var_{U_N}[ \gradFestNt(\lambdavec)] \leq \hat{M}_V \times \frac{1}{\tau^{2t}}
$$
and 
$$
\EE[\F(\lambdavec_{t+1}) - \F(\lambdavec^\star)] \leq \omega \xi^{2t},
$$
where $\omega = \max\{ \alpha L \hat{M}_V /c,  \F(\lambdavec_{1}) - \F(\lambdavec^\star) \}$.
\end{theorem}
This result compares favorably with a standard result on the linear convergence of \textsc{sgd} with fixed step size and strongly convex functions \citep{bottou2016optimization}. For \textsc{mc} sampling one obtains a different constant $\Tilde{\omega}$ and an upper bound with $\xi^{t}$ and not $\xi^{2t}$. Thus, besides the constant factor, \textsc{rqmc} samples allow us to close the optimality gap faster for the same geometric increase in the sample size $\tau^t$ or to use $\tau^{t/2}$ to obtain the same linear rate of convergence as \textsc{mc} based estimators. 

\paragraph{Other Remarks}
The reduced variance in the estimation of the gradients should allow us to make larger moves in the parameter space. This is for example achieved by using adaptive step size algorithms as Adam \citep{kingma2014adam}, or Adagrad \citep{duchi2011adaptive}. However, the theoretical analysis of these algorithms is beyond the scope of this paper. 

Also, note that it is possible to relax the smoothness assumptions on $G$ while supposing only square integrability. 
Then one obtains rates in $\oo(N^{-1})$. Thus, \RQMC{} yields always a faster rate than  \MC{}, regardless of the smoothness. See Appendix~\ref{appendix:info_qmc} for more details. 

In the previous analysis, we have assumed that the entire randomness in the gradient estimator comes from the sampling of the variational distribution. In practice, additional randomness is introduced in the gradient via mini batch sampling. This leads to a dominating term in the variance of $\OO(K^{-1})$ for mini batches of size $K$. Still, the part of the variance related to the variational family itself is reduced and so is the variance of the gradient estimator as a whole. 
\section{Experiments}
\label{sec:experiments}

We study the effectiveness of our method in three different settings: a hierarchical linear regression, a multi-level Poisson generalized linear model (GLM) and a Bayesian neural network (BNN). Finally, we confirm the result of Theorem~\ref{theorem:noise_reduction}, which proposes to  increase the sample size over iterations in \textsc{qmcvi} for faster asymptotic convergence. 

\begin{figure}[h!]
        \includegraphics[width=0.473\textwidth]{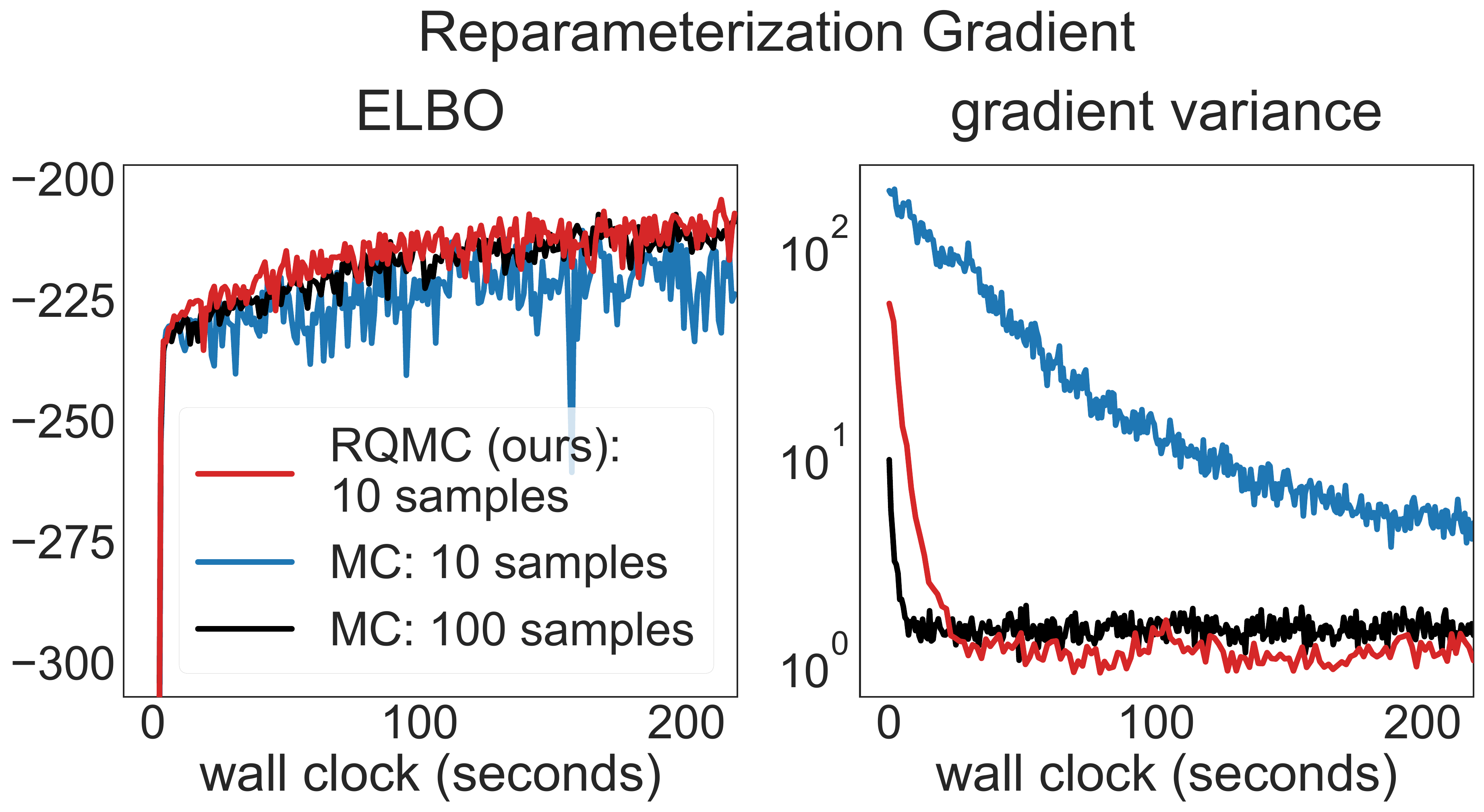}
        \includegraphics[width=0.49\textwidth]{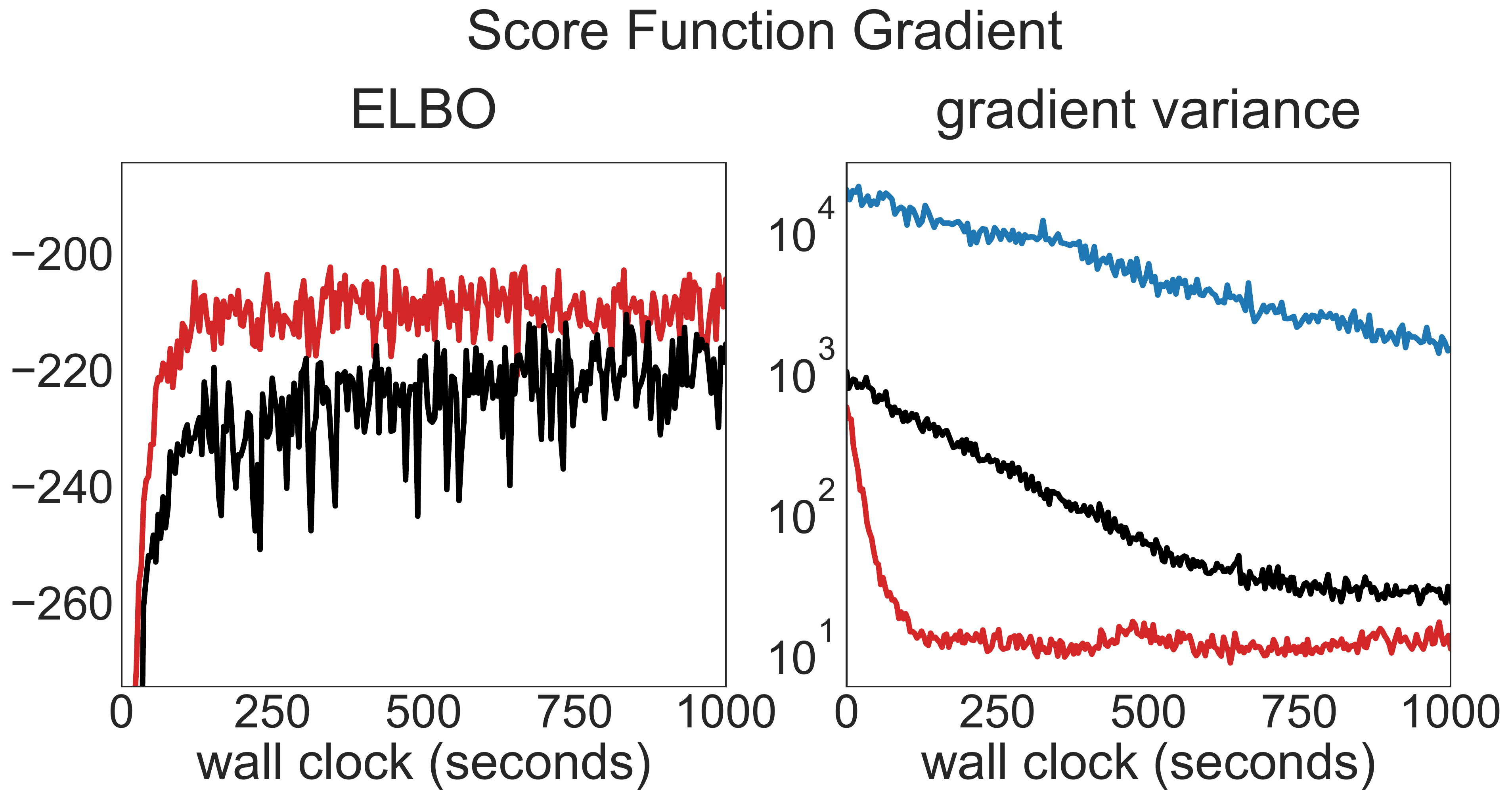}
    \caption[Caption for LOF]{Toy: experiment \ref{sec:exp_toy}. \textsc{elbo} optimization path using Adam and variance of the stochastic gradient using the \textsc{rqmc} based gradient estimator using 10 samples (ours, in red) and the \textsc{mc} based estimator using 10 samples (blue) and 100 samples (black), respectively. The upper panel corresponds to the reparameterization gradient and the lower panel to the score function gradient estimator\protect\footnotemark{}.
    For both versions of \textsc{mcvi}, using \textsc{rqmc} samples (proposed) leads to variance reduction and faster convergence.}
    \label{fig:linear_regression}
\end{figure}
\footnotetext{Using only 10 samples for the \MC{} based score function estimator leads to divergence and the \textsc{elbo} values are out of the scope of the plot.}

\begin{figure*}
    \centering
    \includegraphics[width=0.8\textwidth]{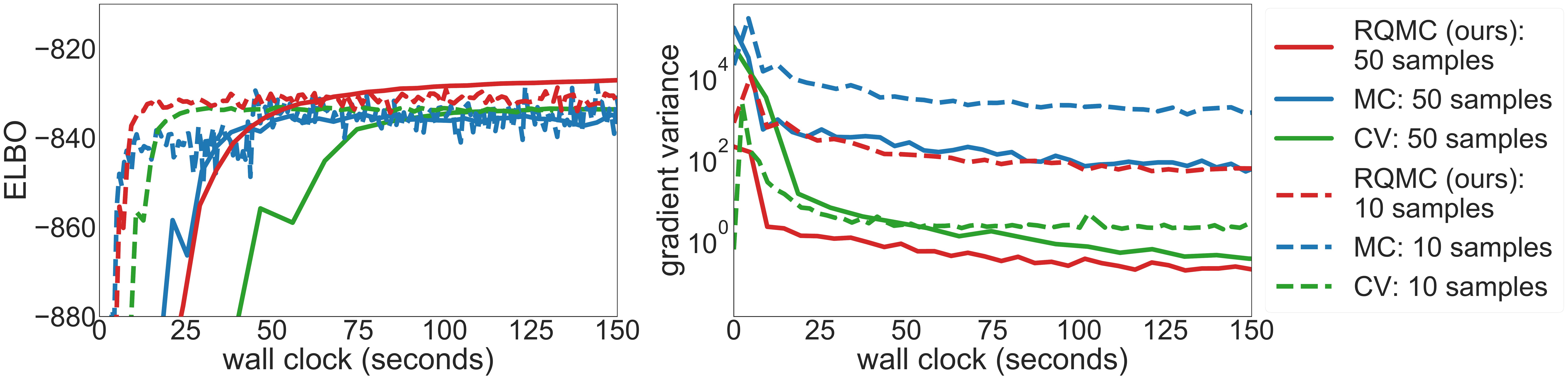}
    \caption{Frisk: experiment \ref{sec:exp_frisk}. Left, the optimization path of the \textsc{elbo} is shown using Adam with the \textsc{rqmc}, \textsc{mc} and \textsc{cv} based reparameterization gradient estimator, respectively. Right, the gradient variances as function of time are reported.
     In the case of using 10 samples (dashed lines) \RQMC{} (ours) outperforms the baselines in terms of speed while the \textsc{cv} based method exhibits lowest gradient variance. When increasing the sample size to 50 (solid lines) for all methods, \RQMC{} converges closer to the optimum than the baselines while having lowest gradient variance.}
    \label{fig:frisk}
%
    \centering
    \includegraphics[width=0.8\textwidth]{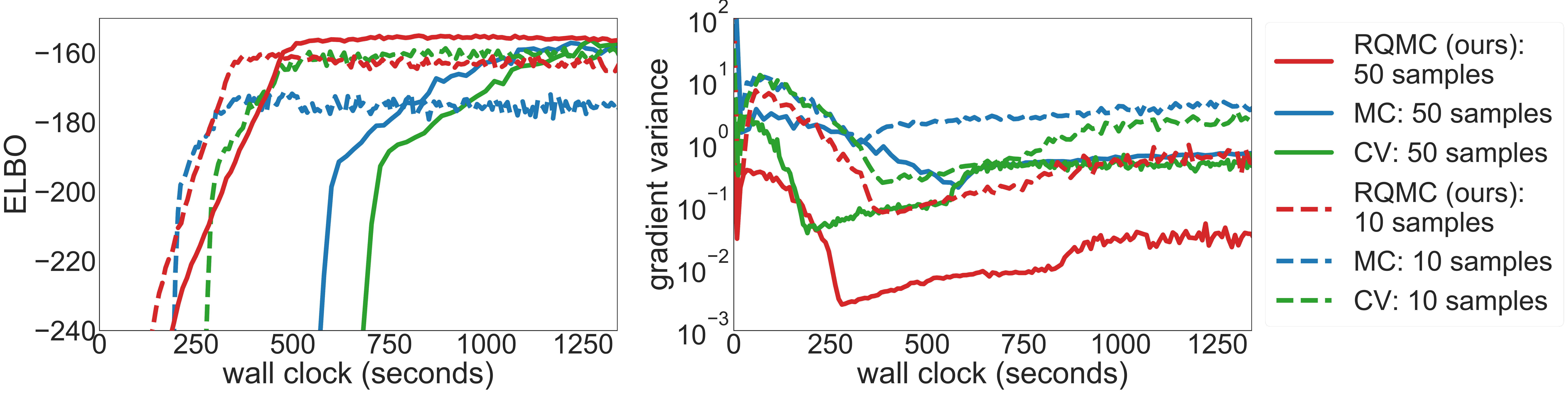}
    \caption{BNN: experiment \ref{sec:exp_bnn}. Left, the optimization path of the \textsc{elbo} is shown using Adam with the \textsc{rqmc}, \textsc{mc} and \textsc{cv} based reparameterization gradient estimator, respectively. Right, the gradient variances as function of time are reported.
    \textsc{rqmc} (ours) based on 10 samples outperforms the  baselines in terms of speed. \textsc{rqmc} with 50 samples is bit slower but converges closer to the optimum as its gradient variance is up to 3 orders of magnitude lower than for the baselines.}
  \label{fig:bnn}
\end{figure*}

\paragraph{Setup}
In the first three experiments we optimize the \EL~using the Adam optimizer~\citep{kingma2014adam} with the initial step size set to $0.1$, unless otherwise stated. The \textsc{rqmc} sequences are generated through a \texttt{python} interface to the \texttt{R} package \texttt{randtoolbox} \citep{randtoolbox}. In particular we use scrambled Sobol sequences. The gradients are calculated using an automatic differentiation toolbox. 
The \EL{} values are computed by using $10,000$ \MC{} samples, the variance of the gradient estimators is estimated by resampling the gradient $1000$ times in each optimization step and computing the empirical variance.

\paragraph{Benchmarks}
The first benchmark is the vanilla \textsc{mcvi} algorithm based on ordinary \textsc{mc} sampling. 
Our method \QMCVI{} replaces the \MC{} samples by \RQMC{} sequences and comes at almost no computational overhead (Section~\ref{sec:method}). 

Our second benchmark in the second and third experiment is the control variate (\textsc{cv}) approach of \citet{DBLP:conf/nips/MillerFDA17}, where we use the code provided with the publication. In the first experiment, this comparison is omitted since the method of \citet{DBLP:conf/nips/MillerFDA17} does not apply in this setting due to the non-Gaussian variational distribution. 

\paragraph{Main Results} We find that our approach generally leads to a faster convergence compared to our baselines due to a decreased gradient variance. For the multi-level Poisson GLM experiment, we also find that our \textsc{rqmc} algorithm converges to a  better local optimum of the \textsc{elbo}. As proposed in Theorem~\ref{theorem:noise_reduction}, we find that increasing the sample size over iteration in \textsc{qmcvi} leads to a better asymptotic convergence rate than in \textsc{mcvi}.


\subsection{Hierarchical Linear Regression}
\label{sec:exp_toy}

We begin the experiments with a toy model of hierarchical linear regression with simulated data. The sampling process for the outputs $y_i$ is 
$y_i~\sim~\mathcal{N}(\xvec_i^\top \mathbf{b}_i,\epsilon), \;  \mathbf{b}_i~\sim~\mathcal{N}(\muvec_\beta,\sigma_\beta).$
We place lognormal hyper priors on the variance of the intercepts $\sigma_\beta$ and on the noise $\epsilon$; and a Gaussian hyper prior on $\muvec_\beta$.
Details on the model are provided in Appendix~\ref{appendix:exp_toy}.
We set the dimension of the data points to be $10$ and simulated $100$ data points from the model. This results in a $1012$-dimensional posterior, which we approximate by a variational distribution that mirrors the prior distributions.

We optimize the \textsc{elbo} using Adam~\citep{kingma2014adam} based on the score function as well as the reparameterization gradient estimator. We compare the standard \textsc{mc} based approach using 10 and $100$ samples with our  \textsc{rqmc} based approach using $10$ samples, respectively. The \CV{} based estimator cannot be used in this setting since it only supports Gaussian variational distributions and the variational family includes a lognormal distribution. 
For the score function estimator, we set the initial step size of Adam to $0.01$. 

The results are shown in Figure~\ref{fig:linear_regression}. We find that using \textsc{rqmc} samples decreases the variance of the gradient estimator substantially. This applies both to the score function and the reparameterization gradient estimator.
Our approach substantially improves the standard score function estimator in terms of convergence speed and leads to a decreased gradient variance of up to three orders of magnitude.
Our approach is also beneficial in the case of the reparameterization gradient estimator, as it allows for reducing the sample size from 100 \MC{} samples to 10 \RQMC{} samples, yielding a similar gradient variance and optimization speed.



\subsection{Multi-level Poisson GLM}
\label{sec:exp_frisk}
We use a multi-level Poisson generalized linear model (GLM), as introduced in \citep{gelman2006data} as an example of multi-level modeling. This model has a $37$-dim posterior, resulting from its  hierarchical structure.  

As in \citep{DBLP:conf/nips/MillerFDA17}, we apply this model to the \textit{frisk} data set \cite{gelman2006analysis} that contains information on the number of stop-and-frisk events within different ethnicity groups.
The generative process of the model is described in Appendix~\ref{appendix:frisk}. We approximate the posterior by a  diagonal Gaussian variational distribution.

The results are shown in Figure~\ref{fig:frisk}.
When using a small number of samples $(N=10)$, all three methods have comparable convergence speed and attain a similar optimum. In this setting, the \CV{} based method has lowest gradient variance. 
When increasing the sample size to $50$, our proposed \RQMC{}  approach leads to substantially decreased gradient variance and allows Adam to convergence closer to the optimum than the baselines. This agrees with the fact that \RQMC{} improves over \MC{} for sufficiently large sample sizes.


\subsection{Bayesian Neural Network}
\label{sec:exp_bnn}

As a third example, we study \QMCVI{} and its baselines in the context of a Bayesian neural network. The network consists of a $50$-unit hidden layer with ReLU activations. We place a normal prior over each weight, and each weight prior has an inverse Gamma hyper prior. We also place an inverse Gamma prior over the observation variance. 
The model exhibits a posterior of dimension $d=653$ and is applied to a 100-row subsample of the wine dataset from the UCI repository\footnote{\url{https://archive.ics.uci.edu/ml/datasets/Wine+Quality}}. The generative process is described in Appendix~\ref{appendix:exp_bnn}. We approximate the posterior by a variational diagonal Gaussian.
%

The results are shown in Figure~\ref{fig:bnn}. For $N=10$, both the \textsc{rqmc} and the \textsc{cv} version converge to a comparable value of the \textsc{elbo}, whereas the ordinary \textsc{mc} approach converges to a lower value. For $N=50$, all three algorithms reach approximately the same value of the \textsc{elbo}, but our \textsc{rqmc} method converges much faster.
In both settings, the variance of the \textsc{rqmc} gradient estimator is one to three orders of magnitude lower than the variance of the baselines.

\subsection{Increasing the Sample Size Over Iterations}
\label{sec:exp_c-sgd}

Along with our new Monte Carlo variational inference approach \textsc{qmcvi},  Theorem~\ref{theorem:noise_reduction} gives rise to a
new stochastic optimization algorithm for Monte Carlo objectives. Here, we investigate this algorithm empirically, using a constant learning rate and an (exponentially) increasing sample size schedule. We show that, for strongly convex objective functions and some mild regularity assumptions, our \textsc{rqmc} based gradient estimator leads to a faster asymptotic convergence rate than using the ordinary \textsc{mc} based gradient estimator. 

 \begin{figure}
    \includegraphics[width=0.46\textwidth]{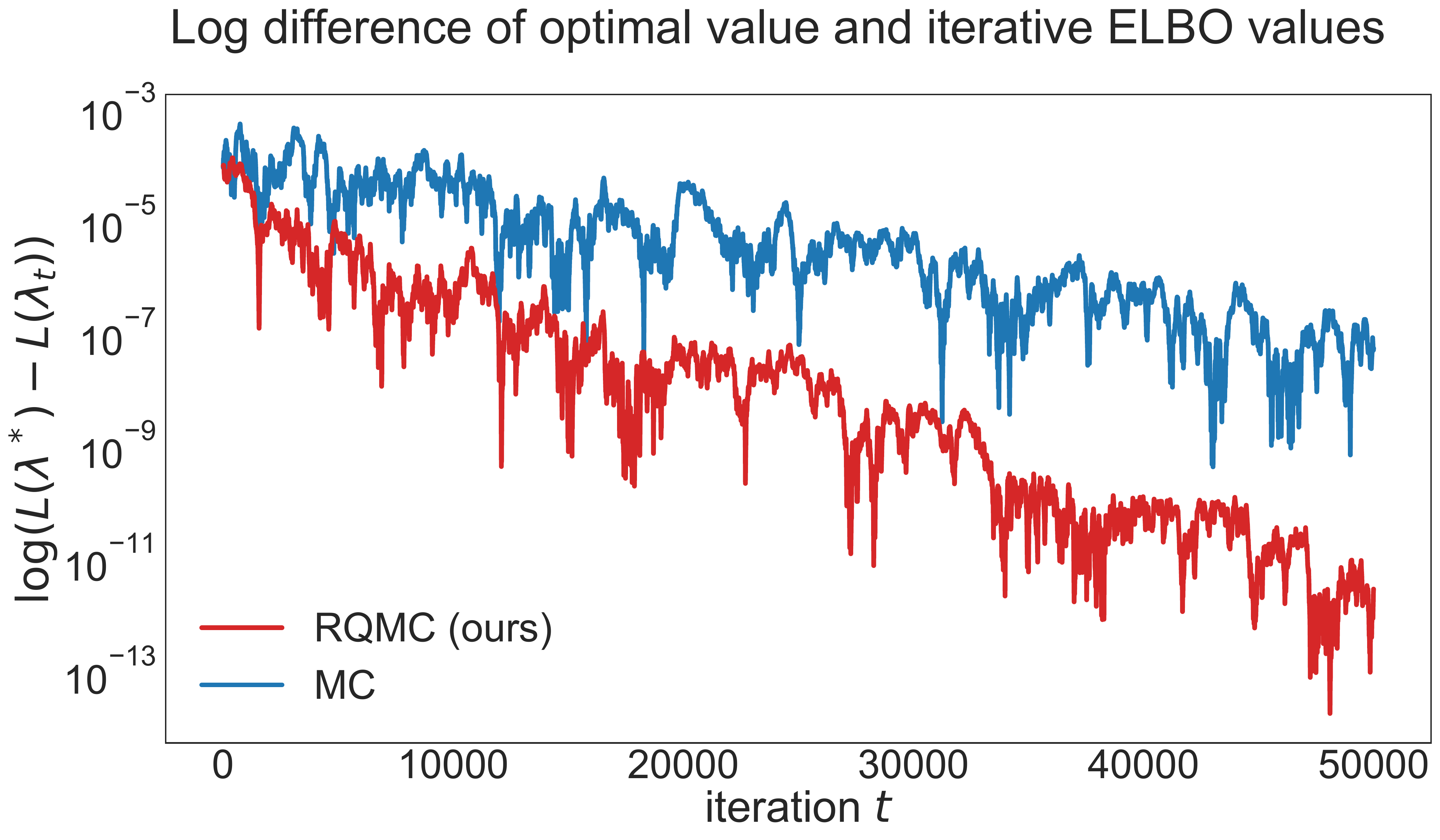}
  	\caption{Constant SGD: experiment \ref{sec:exp_c-sgd}. We exemplify the consequences of Theorem \ref{theorem:noise_reduction} and optimize a simple concave \textsc{elbo} using \textsc{sgd} with fixed learning rate $\alpha=0.001$ while the number of samples are iteratively increased. We use an exponential sample size schedule (starting with one sample and 50.000 samples in the final iteration). The logarithmic difference of the \textsc{elbo} to the optimum is plotted. We empirically confirm the result of Theorem \ref{theorem:noise_reduction} and observe a faster asymptotic convergence rate when using \textsc{rqmc} samples over \textsc{mc} samples.}
  \label{fig:constant_sgd}
\end{figure}

In our experiment, we consider a two-dimensional factorizing normal target distribution with zero mean and standard deviation one. Our variational distribution is also a normal distribution with fixed standard deviation of~$1$, and with a variational mean parameter, i.e., we only optimize the mean parameter. In this simple setting, the \textsc{elbo} is strongly convex and the variational family includes the target distribution. We optimize the \textsc{elbo} with an increasing sample size, using the \textsc{sgd} algorithm described in Theorem~\ref{theorem:noise_reduction}.
We initialize the variational parameter to $(0.1, 0.1)$. Results are shown in Figure~\ref{fig:constant_sgd}. 

We considered both \textsc{rqmc} (red) and \textsc{mc} (blue) based gradient estimators. We plot the difference between the optimal \textsc{elbo} value and the optimization trace in logarithmic scale. The experiment confirms the theoretical result of Theorem~\ref{theorem:noise_reduction} as our \textsc{rqmc} based method attains a faster asymptotic convergence rate than the ordinary \textsc{mc} based approach. This means that, in the absence of additional noise due to data subsampling, optimizing Monte Carlo objectives with \textsc{rqmc} can drastically outperform \textsc{sgd}.
\section{Conclusion} \label{sec:conclusion}

We investigated randomized Quasi-Monte Carlo (\textsc{rqmc}) for stochastic optimization of Monte Carlo objectives. We termed our method Quasi-Monte Carlo Variational Inference (\textsc{qmcvi}), currently focusing on variational inference applications. Using our method, we showed that we can achieve faster convergence due to variance reduction. 

\textsc{qmcvi} has strong theoretical guarantees and provably gets us closer to  the optimum of the stochastic objective. Furthermore, in absence of additional sources of noise such as data subsampling noise, \textsc{qmcvi} converges  at a faster rate than \textsc{sgd} when increasing the sample size over iterations.

\textsc{qmcvi} can be easily integrated into automated inference packages. All one needs to do is replace a sequence of uniform random numbers over the hypercube by an \textsc{rqmc} sequence, and perform the necessary reparameterizations to sample from the target distributions.

An open question remains as to which degree \textsc{qmcvi} can be combined with control variates, as \textsc{rqmc} may introduce additional unwanted correlations between the gradient and the \textsc{cv}. We will leave this aspect for future studies.
We see particular potential for \textsc{qmcvi} in the context of reinforcement learning, which we consider to investigate.

\section*{Acknowledgements} 
We would like to thank Pierre E. Jacob, Nicolas Chopin, Rajesh Ranganath, Jaan Altosaar and Marius Kloft for their valuable feedback on our manuscript. This work was partly funded by the German Research Foundation (DFG) award KL 2698/2-1 and a GENES doctoral research scholarship.

{\small
\bibliographystyle{apa}
\bibliography{bib}

\begin{appendix}
     \section{Additional Information on QMC}
\label{appendix:info_qmc}
We provide some background on \textsc{qmc} sequences that we estimate necessary for the understanding 
of our algorithm and our theoretical results. 
\paragraph{Quasi-Monte Carlo (QMC)}
Low discrepancy sequences (also called Quasi Monte Carlo sequences), are used
to approximate integrals over the $[0,1]^d$ hyper-cube:
$\EE{\psi(U)} = \int_{[0,1]^d} \psi(u) du$,
that is the expectation of the random variable $\psi(U)$, where $ U \sim
\mathcal{U} [0,1]^d$, is a uniform distribution on $[0,1]^d$. The basic Monte Carlo approximation of the
integral is  $\hat{I}_N := \frac{1}{N}
\sumnN \psi(\uvec_n)$, where each $\uvec_n \sim \mathcal{U}[0,1]^d $, independently. 
The error of this approximation is $\OO(N^{-1})$, since 
$\Var[\hat{I}_N] = \Var[\psi(U)]/N$. 

This basic approximation may be improved by replacing the random variables $\uvec_n$ 
by a low-discrepancy sequence; that is, informally, a deterministic sequence that covers 
$[0,1]^d$ more regularly. 
The error of this approximation is assessed by the 
Koksma-Hlawka inequality \citep{doi:10.1002/0471667196.ess4085.pub2}:
\begin{eqnarray} \label{eq:koksma}
\left| \int_{[0,1]^d} \psi(\uvec) d\uvec -  \frac{1}{N} \sumnN \psi(\uvec_n) \right| \leq V(\psi) D^*(\uvec_{1:N}), 
\end{eqnarray}
where $V(\psi)$ is the total variation in the sense of Hardy and Krause \citep{zbMATH02650645}. 
This quantity is closely linked to the smoothness of the function $\psi$.
$D^*(\uvec_{1:N})$ is called the star discrepancy, that measures how well the sequence covers the target space. 

The general notion of discrepancy of a given sequence $\uvec_1, \cdots, \uvec_N$ is defined as follows: 
\begin{eqnarray*}
D(\uvec_{1:N}, \mathcal{A}) := \sup_{A \in \mathcal{A}} \left| \frac{1}{N} \sumnN \1{ \uvec_n \in A }  - \lambda_d(A)\right|,
\end{eqnarray*}
where $\lambda_d(A)$ is the volume (Lebesgue measure on $\mathbb{R}^d$) of $A$ and $\mathcal{A}$ is a set of measurable sets. 
When we fix the sets $A = [0,\mathbf{b}]= \prod_{i=1}^d [0,b_i]$ with $0 \leq b_i \leq 1$ as a products set of intervals anchored at $0$, the star discrepancy is then defined as follows
\begin{eqnarray*}
D^*(\uvec_{1:N}) := \sup_{[0,\mathbf{b}] } \left| \frac{1}{N} \sumnN \1{\uvec_n \in [0,\mathbf{b}] }  - \lambda_d([0,\mathbf{b}])\right|.
\end{eqnarray*}
It is possible to construct sequences such that $D^*(\uvec_{1:N}) = \OO((\log N)^{2d-2}/N^{2})$. See also \citet{kuipers2012uniform} and \citet{leobacher2014introduction} for more details.

Thus, \textsc{qmc} integration schemes are asymptotically more efficient than \textsc{mc}
schemes. However, if the dimension $d$ gets too large, 
the number of necessary samples $N$ in order to reach the asymptotic regime becomes prohibitive. 
As the upper bound is rather pessimistic, in practice \textsc{qmc} integration outperforms \textsc{mc} integration even 
for small $N$ in most applications, see e.g. the examples in Chapter 5 of \citet{glasserman2013monte}.
Popular \textsc{qmc} sequences are for example the Halton sequence or the Sobol sequence. See e.g. \citet{dick_kuo_sloan_2013} for details on the construction. 
A drawback of \textsc{qmc} is that it is difficult 
to assess the error and that the deterministic approximation is inherently biased. 

\paragraph{Ramdomized Quasi Monte Carlo (RQMC).}
The reintroduction of randomness in a low discrepancy sequence while preserving
the low discrepancy properties enables the construction of confidence
intervals by repeated simulation. Moreover, the randomization makes
the approximation unbiased. The simplest method for this purpose is a randomly shifted
sequence. Let $v \sim \mathcal{U}[0,1]^d$. Then the sequence based on
$\hat{\uvec}_n :=  \uvec_n + v \mod 1$
preserves the properties of the \textsc{qmc} sequence with probability $1$
and is marginally uniformly distributed. 

Scrambled nets \citep{owen_scrambled_1997} represent a more sophisticated approach. 
Assuming smoothness of the derivatives of the function, 
\citet{gerber2015integration} showed recently, that rates of $\OO(N^{-2})$  
are achievable. We summarize the best rates in table \ref{tab:rates}.

\begin{table}[]
    \centering
    \begin{tabular}{ccc}
        MC & QMC & RQMC \\
        \hline
        $N^{-1}$ & $N^{-2} (\log N)^{2d-2} $ & $N^{-2}$ 
    \end{tabular}
    \caption{Best achievable rates for \textsc{mc}, \textsc{qmc} and \textsc{rqmc} in terms of the MSE of the approximation.}
    \label{tab:rates}
\end{table}

\paragraph{Transforming \textsc{qmc} and  \textsc{rqmc} sequences}
A generic recipe for using \textsc{qmc} / \textsc{rqmc} for integration is given by transforming a sequence with the inverse Rosenblatt transformation $\Gamma : \uvec \in [0,1]^d \mapsto \zvec \in \RR^d$, see \citet{rosenblatt1952} and \citet{gerber2015sequential}, such that 
$$
\int \psi(\Gamma(\uvec)) d\uvec = \int \psi(\zvec) p(\zvec) d\zvec, 
$$
where $p(\zvec)$ is the respective measure of integration. 
The inverse Rosenblatt transformation can be understood as the multivariate extension of the inverse cdf transform. 
For the procedure to be correct we have to make sure that $\psi \circ \Gamma$ is sufficiently regular. 

\paragraph{Theoretical Results on RQMC}
In our analysis we mainly use the following result. 
\begin{theorem} \label{theorem:owen}
\cite{owen2008local}
Let $\psi:[0,1]^d\rightarrow \mathbb{R}$ be a function such that its cross partial derivatives
up to order $d$ exist and are continuous, 
and let $(\uvec_n)_{n \in 1:N}$ be a relaxed scrambled $(\alpha, s, m, d)$-net in base $b$ with dimension $d$ with
uniformly bounded gain coefficients.
Then,
\[\Var\left(\frac{1}{N} \sumnN \psi(\uvec_n) \right)
	= \mathcal{O}\left(N^{-3} \log(N)^{(d-1)}\right)
	,
\]
where $N=\alpha b^m$.
\end{theorem}
In words, $\forall \tau > 0$ the \textsc{rqmc} error rate is $\OO(N^{-3+\tau})$ when
a scrambled $(\alpha, s, m, d)$-net is used. 
However, a more general result has recently been shown by \citet{gerber2015integration}[Corollary 1], where
if $\psi$ is square integrable and $(\uvec_n)_{n \in 1:N}$ is a scrambled $(s, d)$-sequence, then 
	\[
	\Var\left(\frac{1}{N} \sumnN \psi(\uvec_n) \right)
	= \oo \left(N^{-1}\right).
	\]
This result shows that \textsc{rqmc} integration is always better than MC integration. 
Moreover, \citet{gerber2015integration}[Proposition 1] shows that rates $\OO(N^{-2})$ can be obtained 
when the function $\psi$ is regular in the sense of Theorem \ref{theorem:owen} . In particular one gets 
	\[
	\Var\left(\frac{1}{N} \sumnN \psi(\uvec_n) \right)
	= \OO \left(N^{-2}\right).
	\]

\section{Proofs}
Our proof are deduced from standard results in the stochastic approximation literature, e.g. \citet{bottou2016optimization},
when the variance of the gradient estimator is reduced due to \textsc{rqmc} sampling. Our proofs rely on scrambled $(s, d)$-sequences in order to use the result of \citet{gerber2015integration}. The scrambled Sobol sequence, that we use in our simulations satisfies the required properties. 
We denote by $\EE$ the total expectation and by $\EEUNt$ the expectation with respect to the \textsc{rqmc} sequence $U_N$ generated at time $t$. Note, that $\lambdavect$ is not a random variable w.r.t. $U_{N,t}$ as it only depends on all the previous 
$U_{N,1}, \cdots, U_{N,t-1}$ due to the update equation in \eqref{eq:sgd_update}. However, $\FestN (\lambdavect )$ is a random variable depending on $\UNt$. 

\subsection{Proof of Theorem \ref{theorem:lipschitz_functions}}
Let us first prove that $\tr \Varr{ \gradFestN(\lambdavec) } \leq M_V \times r(N)$ for all $\lambdavec$. 
By assumption we have that 
$g_z(\lambdavec)$ with $z=\Gamma(\uvec)$ is a function $G: \uvec \mapsto g_{\Gamma(\uvec)}(\lambdavec)$ with continuous mixed partial derivatives of up to order $d$ for all $\lambdavec$. Therefore, if $\uvec_1, \cdots, \uvec_N$ is a \textsc{rqmc} sequence, then the trace of the variance of the estimator $\gradFestN(\lambdavec)$ is upper bounded by Theorem \ref{theorem:owen} and its extension by \citet{gerber2015integration}[Proposition 1] by a uniform bound $M_V$ and the quantity $r(N) = \OO(N^{-2})$, that goes to $0$ faster than the Monte Carlo rate $1/N$. 

By the Lipschitz assumption we have that 
$\F(\lambdavec) \leq \F(\lambdaveco) + \nabla \F(\lambdaveco)^T(\lambdavec - \lambdaveco) + \frac{1}{2}L \norm{ \lambdavec - \lambdaveco} $, $\forall \lambdavec, \lambdaveco$, see for example \citet{bottou2016optimization}. 
By using the fact that $\lambdavectone-\lambdavect = - \alpha \gradFestN (\lambdavect )$ we obtain
\begin{eqnarray*}
&& \F(\lambdavectone)-\F(\lambdavect) \\
&\leq& \nabla \F(\lambdavect)^T(\lambdavec_{t+1}-\lambdavect)  + \frac{1}{2}L \norm{ \lambdavectone-\lambdavect }, \\
&=& - \alpha \nabla \F(\lambdavect)^T\gradFestN (\lambdavect) +  \frac{\alpha^2 L}{2} \norm{ \gradFestN(\lambdavect) } .
\end{eqnarray*}
After taking expectations with respect to $\UNt$ we obtain 
\begin{eqnarray*}
&& \EEUNt \F(\lambdavectone)- \F(\lambdavect) \\
&\leq& - \alpha \nabla \F(\lambdavect) \EEUNt \gradFestN(\lambdavect)  +  \frac{\alpha^2 L}{2} \EEUNt \norm{ \gradFestN(\lambdavect) }.
\end{eqnarray*}
We now use the fact that $\EEUNt \norm{ \gradFestN(\lambdavect)} = \tr \VarrUNt{ \gradFestN(\lambdavect)} +  \norm{ \EEUNt  \gradFestN(\lambdavect) }$ and after exploiting the fact that $\EEUNt  \gradFestN(\lambdavect) = \nabla \F(\lambdavect)$ we obtain 
\begin{eqnarray*} 
&& \EEUNt \F(\lambdavectone)- \F(\lambdavect) \\
&\leq& - \alpha \norm{ \nabla \F(\lambdavect) }   +  \frac{\alpha^2 L}{2} \left[ \tr \VarrUNt{ \gradFestN(\lambdavect)} 
 +  \norm{  \nabla \F(\lambdavect) } \right], \\
&=&   \frac{\alpha^2 L}{2} \tr \VarrUNt{ \gradFestN(\lambdavect)} + \left( \frac{ \alpha^2L}{2} - \alpha \right) \norm{  \nabla \F(\lambdavect) }. \label{eq:inequality_proof}
\end{eqnarray*}
The inequality is now summed for $t=1,\cdots,T$ and we take the total expectation:
\begin{eqnarray*}
&& \EE \F(\lambdavec_T)- \F(\lambdavec_1) \\
&\leq&   \frac{\alpha^2 L}{2} \sum_{t=1}^T \EE \tr \VarrUNt{ \gradFestN(\lambdavect)} \\
&& +  \left(\frac{\alpha^2 L}{2} - \alpha \right) \sumtT \EE \norm{  \nabla \F(\lambdavect) } .
\end{eqnarray*}
We use the fact that 
$ \F(\lambdavec^\star) - \F(\lambdavec_1) \leq \EE \F(\lambdavec_T) - \F(\lambdavec_1)$, where $\lambdavec_1$ is deterministic and $\lambdavec^\star$ is the true minimizer, and divide the inequality by $T$:
\begin{eqnarray*}
&& \frac{1}{T} \left[ \F(\lambdavec^\star)- \F(\lambdavec_1) \right] \\
&\leq&   \frac{\alpha^2L}{2} \frac{1}{T} \sum_{t=1}^T \EE \tr \VarrUNt{ \gradFestN(\lambdavect) } \\
&& + \left( \frac{\alpha^2L}{2} - \alpha\right) \frac{1}{T} \sumtT \EE \norm{  \nabla \F(\lambdavect) } .
\end{eqnarray*}
By rearranging and using $\alpha < 2/L$ and $\mu = 1 - \alpha L/2$ we obtain 
\begin{eqnarray*}
&& \frac{1}{T} \sum_{t=1}^T \EE \norm{ \nabla \F(\lambdavect) } \\
&\leq& \frac{1}{T \alpha \mu} \left[ \F(\lambdavec_1)- \F(\lambdavec^\star) \right] \\
&&+  \frac{\alpha L}{2 \mu} \frac{1}{T} \sum_{t=1}^T \EE \tr \VarrUNt{ \gradFestN(\lambdavect) }.
\end{eqnarray*}
We now use $\tr \VarrUNt{ \gradFestN(\lambdavect) } \leq M_V r(N)$ for all $t$. 
Equation \eqref{eq:limit_lipschitz} is obtained as $T \rightarrow \infty$.

\subsection{Proof of Theorem \ref{theorem:strongly_convex_functions}}
A direct consequence of strong convexity is the fact that the optimality gap can be upper bounded by the gradient in the current point $\lambdavec$, e.g. 
$2c(\F(\lambdavec) - \F(\lambdavec^\star)) \leq \norm{ \nabla \F(\lambdavec)}, \forall \lambdavec$. The following proof uses this result. 
Based on the previous proof we get 
\begin{eqnarray*}
&& \EEUNt \F(\lambdavectone)- \F(\lambdavect) \\
&\leq&  \frac{\alpha^2 L}{2} \tr \VarrUNt{ \gradFestN(\lambdavect)} \\
&& + \left( \frac{\alpha^2 L}{2} - \alpha \right) \norm{  \nabla \F(\lambdavect) } \\
&\leq&  \frac{\alpha^2 L}{2} \tr \VarrUNt{ \gradFestN(\lambdavect)} \\
&& + \left( \frac{\alpha^2 L}{2} - \alpha \right) 2c \left( \F(\lambdavect) - \F(\lambdavec^\star) \right),
\end{eqnarray*}
where we have used that $\left(\frac{\alpha L}{2} -1 \right) \leq 0$. By subtracting $\F(\lambdavec^\star)$ from both sides, taking total expectations and rearranging we obtain: 
\begin{eqnarray*}
&& \EE \F(\lambdavectone)- \F(\lambdavec^\star) \\ 
&\leq&  \frac{\alpha^2 L}{2} \EE \tr \VarrUNt{\gradFestN(\lambdavect)} \\
&&+ \left[ \left( \frac{\alpha^2 L}{2} - \alpha \right) 2c +1 \right ]\left( \EE \F(\lambdavect) - \F(\lambdavec^\star) \right). 
\end{eqnarray*}
Define $\beta = \left[ \left( \frac{\alpha^2 L}{2} - \alpha \right) 2c +1 \right ]$.
We add 
$$
\frac{\alpha^2 L \EE \tr \VarrUNt{ \gradFestN(\lambdavect)}}{2(\beta -1)}
$$
to both sides of the equation. This yields

\begin{eqnarray*}
&& \EE \F(\lambdavectone)- \F(\lambdavec^\star) 
+ \frac{\alpha^2 L \EE \tr \VarrUNt{ \gradFestN(\lambdavect)} }{2(\beta -1)} \\
&\leq&  \frac{\alpha^2 L}{2} \EE \tr \VarrUNt{ \gradFestN(\lambdavect) }\\ 
&& +  \beta \left( \EE \F(\lambdavect) - \F(\lambdavec^\star) \right) +\frac{\alpha^2 L \EE \tr \VarrUNt{ \gradFestN(\lambdavect)}}{2(\beta -1)} \\
&\leq& \beta \left( \EE \F(\lambdavect) - \F(\lambdavec^\star)  
 + \frac{\alpha^2 L \EE \tr \VarrUNt{ \gradFestN(\lambdavect)}}{2(\beta -1)} \right).
\end{eqnarray*}
Let us now show that $\beta < 1$:
\begin{eqnarray*}
\beta  &\leq& \left[ \left( \frac{\alpha L}{2} -1 \right) 2\alpha c +1 \right ]
\end{eqnarray*}
And as $\frac{\alpha L}{2} < 1$ we get $\beta < 1 -  2\alpha c$. Using $\alpha < 1/2c$ we obtain 
$\beta < 1$ and thus get a contracting equation when iterating over $t$:
\begin{eqnarray*}
&& \EE \F(\lambdavectone)- \F(\lambdavec^\star) \\
&\leq& \beta^t \left( \F(\lambdavec_1) - \F(\lambdavec^\star)  + \frac{\alpha^2 L \EE \tr \VarrUNt{ \gradFestN(\lambdavect)}}{2(\beta -1)} \right) \\
&& -\frac{\alpha^2 L \EE \tr \VarrUNt{ \gradFestN(\lambdavect)}}{2(\beta -1)}, \\
&\leq &
\beta^t \left( \F(\lambdavec_1) - \F(\lambdavec^\star)  + \frac{ \alpha^2 L M_V r(N)}{2(\beta -1)} \right)  \\
&& + \frac{ \alpha^2 L M_V r(N)}{2(1 - \beta)}.
\end{eqnarray*}
After simplification we get 

\begin{eqnarray*}
&& \EE \F(\lambdavectone)- \F(\lambdavec^\star) \\
&\leq& \beta^t \left( \F(\lambdavec_1) - \F(\lambdavec^\star)  + \frac{\alpha L }{2 \alpha L c - 4c} M_V r(N) \right) \\
&& + \frac{\alpha L }{4c - 2 \alpha L c} M_V r(N), 
\end{eqnarray*}

where the first term of the r.h.s. goes to $0$ as $t \rightarrow \infty$.

\subsection{Proof of Theorem \ref{theorem:noise_reduction}}
\label{appendix:thm3}
We require $\underline{N} \geq b^{s+d}$, due to a remark of \citet{gerber2015integration}, where $d$ is the dimension and $b,s$ are integer parameters of the \textsc{rqmc} sequence. 
As $\uvec \mapsto g_{\Gamma(\uvec)}(\lambdavec)$ has continuous mixed partial derivatives of order $d$ for all $\lambdavec$, 
$\tr \Varr{\gradFestNt(\lambdavec)} = \OO(1/N_t^2)$ and consequently 
$\tr \Varr{\gradFestNt(\lambdavec)} \leq \hat{M}_V \times (1/N_t^2)$, where $\hat{M}_V$ is an universal upper bound on the variance. We recall that 
$N_t = \underline{N} + \lceil \tau^{t} \rceil$. 
Consequently 
$$\tr \Varr{\gradFestNt(\lambdavec)} \leq \hat{M}_V \times \frac{1}{(\underline{N} + \lceil \tau^{t} \rceil)^2} \leq  \hat{M}_V \times \frac{1}{ \tau^{2t}}. $$
Now we take an intermediate result from the previous proof: 
\begin{eqnarray*}
&& \EEUNt \F(\lambdavectone)- \F(\lambdavect) \\
&\leq&  \frac{\alpha^2 L}{2} \tr \VarrUNt{ \gradFestNt(\lambdavect) }
+ \left( \frac{\alpha^2 L}{2} - \alpha \right) \norm{  \nabla \F(\lambdavect) } \\
&\leq&  \frac{\alpha^2 L}{2} \tr \VarrUNt{ \gradFestNt(\lambdavect)} 
-  \frac{\alpha}{2}\norm{  \nabla \F(\lambdavect) } \\
&\leq&  \frac{\alpha^2 L}{2} \tr \VarrUNt{\gradFestNt(\lambdavect)} - \alpha c \left( \F(\lambdavect) - \F(\lambdavec^\star) \right),  
\end{eqnarray*}
where we have used $\alpha \leq \min \{1/c, 1/L\}$ as well as strong convexity.
Adding $\F(\lambdavec^\star)$, rearranging and taking total expectations yields
\begin{eqnarray*}
&& \EE \F(\lambdavectone)- \F(\lambdavec^\star) \\
&\leq&  \frac{\alpha^2 L}{2} \EE \tr \VarrUNt{\gradFestNt(\lambdavect)}\\
&&+ [1 - \alpha c] \left( \EE \F(\lambdavect) - \F(\lambdavec^\star) \right). 
\end{eqnarray*}
We now use $\tr \VarrUNt{\gradFestNt(\lambdavect)} \leq \hat{M}_V \xi^{2t}$ and get 
\begin{eqnarray*}
&& \EE \F(\lambdavectone)- \F(\lambdavec^\star) \\
&\leq&  \frac{\alpha^2 L}{2} \hat{M}_V \xi^{2t} + [1 - \alpha c] \left( \EE \F(\lambdavect) - \F(\lambdavec^\star) \right). 
\end{eqnarray*}

We now use induction to prove the main result. The initialization for $t=0$ holds true by the definition of $\omega$. 
Then, for all $t\geq 1$, 
\begin{eqnarray*}
&& \EE \F(\lambdavectone)- \F(\lambdavec^\star) \\
&\leq& [1 - \alpha c] \omega \xi^{2t} + \frac{\alpha^2 L}{2} \hat{M}_V \xi^{2t}, \\
&\leq& \omega \xi^{2t} \left(1 - \alpha c  + \frac{\alpha^2 L \hat{M}_V}{2 \omega} \right), \\
&\leq& \omega \xi^{2t} \left(1 - \alpha c  + \frac{\alpha c}{2} \right), \\
&\leq& \omega \xi^{2t} \left(1 - \frac{\alpha c}{2} \right) \\
&\leq& \omega \xi^{2(t+1)}, 
\end{eqnarray*}
where we have used the definition of $\omega$ and that $\left(1 - \frac{\alpha c}{2} \right) \leq \xi^2$.

\section{Details for the Models Considered in the Experiments}
 
\subsection{Hierarchical Linear Regression}
\label{appendix:exp_toy}
The generative process of the hierarchical linear regression model is as follows.
%
%
\begin{alignat*}{2}
&\muvec_\beta 		\sim \mathcal{N}(0,10^2) 	&&\text{intercept hyper prior}\\
&\sigma_\beta 		\sim \text{LogNormal(0.5)} \hspace{1cm}	&&\text{intercept hyper prior}\\
&\epsilon 		\sim \text{LogNormal(0.5)} 	&&\text{noise}\\
&\mathbf{b}_i		\sim \mathcal{N}(\muvec_\beta,\sigma_\beta) 	&&\text{intercepts}\\
&y_i		\sim \mathcal{N}(\xvec_i^\top \mathbf{b}_i,\epsilon) 	&&\text{output}\\
\end{alignat*}
The dimension of the parameter space is 
$d=I\times k + k+2$, where $k$ denotes the dimension of the data points $x_i$ and $I$ their number. We set $I=100$ and $k=10$. The dimension hence equals $d=1012$. 


\subsection{Multi-level Poisson GLM}
\label{appendix:frisk}
The generative process of the multi-level Poisson GLM is
%
%
%
\begin{alignat*}{2}
&\mu 		\sim \mathcal{N}(0,10^2) 	&&\text{mean offset}\\
&\log \sigma_\alpha^2,\, \log \sigma_\beta^2 	\sim \mathcal{N}(0,10^2)	&&\text{group variances}\\
&\alpha_e 	\sim \mathcal{N}(0,\sigma_\alpha^2)	&&\text{ethnicity effect}\\
&\beta_p 	\sim \mathcal{N}(0,\sigma_\beta^2)		&&\text{precinct effect}\\
&\log \lambda_{ep}	= \mu + \alpha_e + \beta_p  + \log N_{ep}	\hspace{0.2cm}	&&\text{log rate}\\
&Y_{ep}		\sim \text{Poisson}(\lambda_{ep})	&&\text{stop-and-frisk events}
\end{alignat*}
$Y_{ep}$ denotes the number of stop-and-frisk events within ethnicity group $e$ and precinct $p$ over some fixed period. $N_{ep}$ represents the total number of arrests of  group $e$ in precinct $p$ over the same period; $\alpha_e$ and $\beta_p$ are the ethnicity and precinct effects.

\subsection{Bayesian Neural Network}
\label{appendix:exp_bnn}

We study a Bayesian neural network which consists of a 50-unit hidden layer with ReLU activations. 

The generative process is
%
%
\begin{alignat*}{2}
&\alpha		\sim \text{InvGamma} (1,0.1)	&&\text{weight hyper prior}\\
&\tau	\sim \text{InvGamma} (1,0.1)	&&\text{noise hyper prior}\\
&w_i	\sim \mathcal{N}(0,1/\alpha)	&&\text{weights}\\
&y 		\sim \mathcal{N}(\phi(\xvec,\mathbf{w}),1/\tau)	\hspace{1.5cm} &&\text{output distribution}
\end{alignat*}
Above, $w$ is the set of weights, and $\phi(\xvec,\mathbf{w})$ is a multi-layer perceptron that maps input $\xvec$ to output $y$ as a function of parameters $\mathbf{w}$. We denote the set of parameters as $\mathbf{\theta}:= (\mathbf{w},\alpha,\tau)$. The model exhibits a posterior of dimension $d=653$.

%

\section{Practical Advice for Implementing QMCVI in Your Code}
\label{appendix:rqmc_practice}
\renewcommand{\lstlistingname}{Code}

It is easy to implement \RQMC{} based stochastic optimization in your existing code. First, you have to look for all places where random samples are used. Then replace the ordinary random number generator by \RQMC{} sampling. To replace an ordinarily sampled random variable $\zvec$ by an \RQMC{} sample we need a mapping $\Gamma$ from a uniformly distributed random variable $\uvec$ to $\zvec  = \Gamma(\uvec | \lambdavec)$, where $\lambdavec$ are the parameters of the distribution (e.g. mean and covariance parameter for a Gaussian random variable). Fortunately, such a mapping can often be found (see Appendix~\ref{appendix:info_qmc}).

In many recent machine learning models, such as variational auto encoders \citep{kingma2014adam} or generative adversarial networks \citep{goodfellow2014generative}, the application of \RQMC{} sampling is straightforward. In those models, all random variables are often expressed as transformations of Gaussian random variables via deep neural networks. To apply our proposed \RQMC{} sampling approach we only have to replace the Gaussian random variables of the base distributions by \RQMC{} sampled random variables.

In the following Python code snippet we show how to apply our proposed \RQMC{} sampling approach in such settings.

\begin{lstlisting}[language=Python, caption=Python code for \RQMC{} sampling from a Gaussian distribution.]
import numpy.random as nr
import numpy as np
import rpy2.robjects.packages as rpackages
import rpy2.robjects as robjects
from scipy.stats import norm


randtoolbox = rpackages.importr('randtoolbox')


def random_sequence_rqmc(dim, i=0, n=1, random_seed=0):
    """
    generate uniform RQMC random sequence
    """
    dim = np.int(dim)
    n = np.int(n)
    u = np.array(randtoolbox.sobol(n=n, dim=dim, init=(i==0), scrambling=1, seed=random_seed)).reshape((n,dim))
    # randtoolbox for sobol sequence
    return(u)

def random_sequence_mc(dim, n=1, random_seed=0):
    """
    generate uniform MC random sequence
    """
    dim = np.int(dim)
    n = np.int(n)
    np.random.seed(seed=random_seed)
    u = np.asarray(nr.uniform(size=dim*n).reshape((n,dim)))
    return(u)

def transfrom_uniform_to_normal(u, mu, sigma):
    """
    generat a multivariate normal based on 
    a unifrom sequence
    """
    l_cholesky = np.linalg.cholesky(sigma)
    epsilon = norm.ppf(u).transpose()
    res = np.transpose(l_cholesky.dot(epsilon))+mu
    return res

if __name__ == '__main__':
    # example in dimension 2
    dim = 2
    n = 100
    mu = np.ones(dim)*2. # mean of the Gaussian
    sigma = np.array([[2.,1.],[1.,2.]]) # variance of the Gaussian
    
    # generate Gaussian random variables via RQMC
    u_random = random_sequence_rqmc(dim, i=0, n=n, random_seed=1)
    x_normal = transfrom_uniform_to_normal(u_random, mu, sigma)
    
    # here comes the existing code of your model
    deep_bayesian_model(x_normal)

\end{lstlisting}

\end{appendix}
}

\end{document}